\documentclass[10pt,letterpaper,twocolumn]{article}
\usepackage[latin9]{inputenc}
\usepackage{array}
\usepackage{amstext}
\usepackage{graphicx}
\PassOptionsToPackage{normalem}{ulem}
\usepackage{ulem}
\usepackage[unicode=true,
 bookmarks=false,
 breaklinks=false,pdfborder={0 0 1},backref=page,colorlinks=false]
 {hyperref}
\hypersetup{
 pagebackref,breaklinks,colorlinks,allcolors=iccvblue}

\makeatletter

\providecommand{\tabularnewline}{\\}

\usepackage{iccv}              

\definecolor{iccvblue}{rgb}{0.21,0.49,0.74}

\def\paperID{*****} 
\def\confName{ICCV}
\def\confYear{2025}
\usepackage{colortbl}

\author{
Tuyen Tran \quad Thao Minh Le \quad Truyen Tran \\
Applied Artificial Intelligence Institute, Deakin University, Australia \\
{\tt\small \{t.tran,thao.le,truyen.tran\}@deakin.edu.au}
}

\makeatother

\begin{document}
\title{Towards Agentic AI for Multimodal-Guided Video Object Segmentation}

\maketitle
\maketitle 

\global\long\def\ModelName{\text{\emph{M}}^{2}\text{\emph{-Agent}}}%

\global\long\def\Task{\text{Ref-AVS}}%

\begin{abstract}
Referring-based Video Object Segmentation is a multimodal problem
that requires producing fine-grained segmentation results guided by
external cues. Traditional approaches to this task typically involve
training specialized models, which come with high computational complexity
and manual annotation effort. Recent advances in vision-language foundation
models open a promising direction toward training-free approaches.
Several studies have explored leveraging these general-purpose models
for fine-grained segmentation, achieving performance comparable to
that of fully supervised, task-specific models. However, existing
methods rely on fixed pipelines that lack the flexibility needed to
adapt to the dynamic nature of the task. To address this limitation,
we propose \textbf{\uline{M}}ulti-\textbf{\uline{M}}odal Agent
($\ModelName$), a novel agentic system designed to solve this task
in a more flexible and adaptive manner. Specifically, $\ModelName$
leverages the reasoning capabilities of large language models (LLMs)
to generate dynamic workflows tailored to each input. This adaptive
procedure iteratively interacts with a set of specialized tools designed
for low-level tasks across different modalities to identify the target
object described by the multimodal cues. Our agentic approach demonstrates
clear improvements over prior methods on two multimodal-conditioned
VOS tasks: RVOS and Ref-AVS.

\end{abstract}

\section{Introduction \label{sec:intro}}

\begin{figure}[t]
\centering{}\includegraphics[width=0.99\columnwidth]{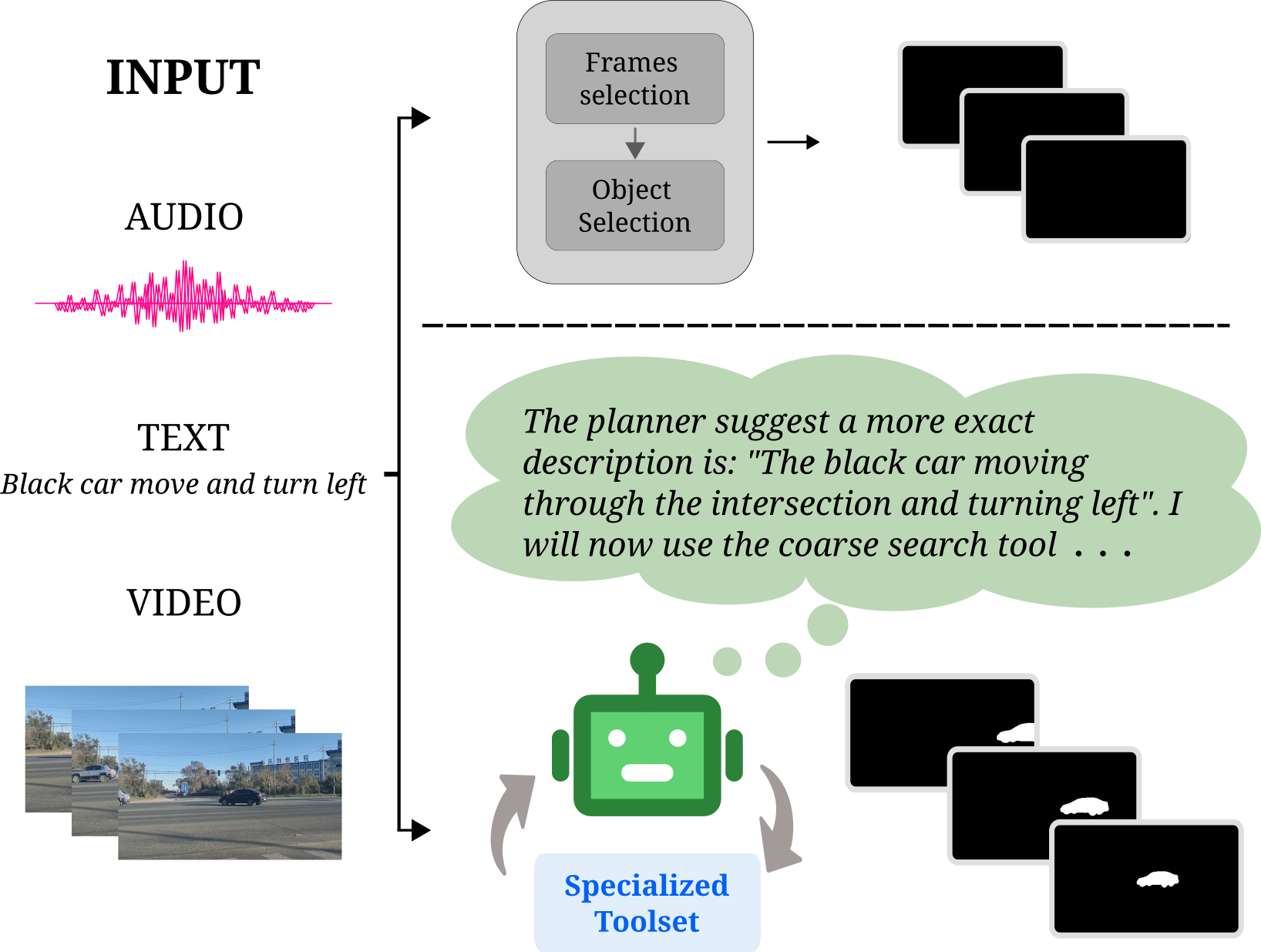}\caption{A comparison between the proposed $\protect\ModelName$ (bottom right)
and baseline AL-Ref-SAM 2 \cite{huang2025unleashing} (top right).
The baseline follows a fixed two-step pipeline composed of frame selection
and object selection, regardless of scenario complexity. In contrast,
our proposed approach leverages the reasoning capabilities of LLMs
to construct a case-specific reasoning chain. $\protect\ModelName$
dynamically decomposes the task into a variable number of steps and
iteratively interacts with a specialized toolset in each step, resulting
in optimal results for each situation. \label{fig:teaser}}
\end{figure}

Video Object Segmentation (VOS) aims at segmenting specific objects
across all frames of a video. This is a fundamental problem in video
understanding with a wide range of real-world applications. Recent
efforts have extended this task by incorporating additional modalities,
including language in Referring Video Object Segmentation \cite{MeViS}
(RVOS) and audio in Reference Audio-Visual Segmentation \cite{wang2024ref}
(Ref-AVS), to enhance its relevance to real-world scenarios. These
task extensions relax the constraint of pre-defined object categories,
allowing the recognition of arbitrary instances described through
multimodal cues. Despite improvement in real-world utility, these
tasks introduce additional complexity in handling multimodal information.

Traditional approaches rely on fully supervised models, requiring
heavy computation and pixel-level annotations. Also, while achieving
strong benchmark performance, these task-specific models often struggle
to generalize beyond their training domains to diverse real-world
scenarios. The emergence of large multimodal foundation models has
enabled training-free approaches to multimodal VOS. Ren et al. \cite{ren2024grounded}
pioneer this direction with a pipeline using GroundingDINO \cite{liu2024grounding}
for first-frame object detection and SAM2 \cite{ravi2024sam} for
video segmentation. However, GroundingDINO, designed for spatial grounding,
struggles with spatiotemporal references. Also, the assumption that
the target object always appears in the first frame does not hold
in real-world scenarios. AL-Ref-SAM 2 \cite{huang2025unleashing}
partially addresses these issues by leveraging GPT's spatiotemporal
reasoning to construct a two-stage pipeline: it first selects an optimal
pivot frame from a fixed set, then identifies the bounding box that
best captures the target instance. While AL-Ref-SAM 2 improves upon
earlier methods, its fixed pipeline remains suboptimal for diverse
scenarios. As shown in Figure \ref{fig:teaser}, the target event
only appears in a specific short duration over entire long video and
cannot be detected from fixed candidate frame list, resulting in incorrect
results. Also, rigid pipeline in AL-Ref-SAM 2 is single-pass without
feedback loop, so failure in one step leads to overall failure.

To bridge this gap, we propose a flexible approach for VOS tasks that
involves query-specific planning and multi-step reasoning through
interaction with specialized vision tools. The approach begins with
a Planner module that generates a tailored execution plan for each
input. This plan then initiates a multi-step reasoning process, where
a large language model dynamically interacts with a suite of audio-vision
tools to resolve the target object. Rather than following a fixed,
hand-crafted pipeline, the system adapts its behavior on the fly based
on contextual situation. This adaptive framework forms \textbf{\uline{M}}ulti-\textbf{\uline{M}}odal
Agent ($\ModelName$), a multimodal AI agent capable of perceiving
diverse inputs to solve the VOS problem. We validate the effectiveness
of $\ModelName$ on two tasks including RVOS and $\Task$. The superior
performance over prior methods on these tasks highlights the effectiveness
of the proposed agentic workflow for this multimodal problem.

\section{Method\label{sec:Method}}

\begin{figure*}[t]
\centering{}\includegraphics[width=0.92\textwidth]{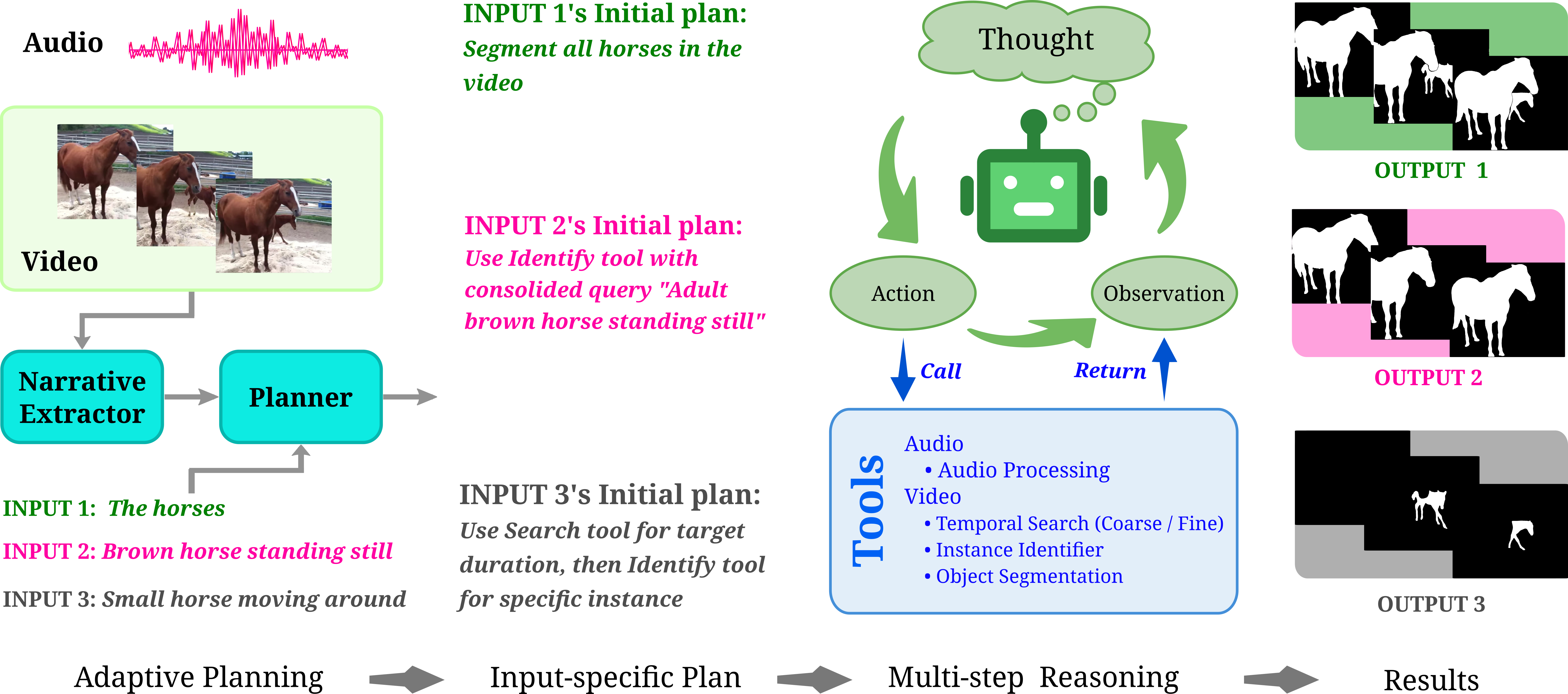}\caption{Given a video and multimodal cues including a textual description
and optionally audio, the Multi-Modal Agent ($\protect\ModelName$)
performs case-specific planning and reasoning to process the inputs.
The approach begins by extracting a textual narrative from the video.
Based on the narrative, the Planner generates an initial tailored
plan for each input query, which initiates a multi-step reasoning
process to identify the referred instance. This reasoning path is
dynamically constructed on the fly through interactions between the
LLM and a predefined set of tools. Audio cues are handled by an audio
processing tool to identify sound sources, while the video is analyzed
by specialized vision tools to locate the target object, resulting
in an optimal reasoning trace tailored to each case. \label{fig:method} }
\end{figure*}

\subsection{Planner }

Real-world visual understanding tasks present highly variable scenarios
and contexts, requiring AI systems to dynamically adapt with case-specific
solutions. In $\ModelName$, a text-based LLM acts as a Planner, generating
tailored plans based on the video content and input query. Specifically,
we extract a detailed textual narrative that captures the object's
progression throughout the video. This narrative is generated once
using a vision-language foundation model, referred to as the Narrative
Extractor. As shown in Figure \ref{fig:method}, the Planner leverages
the narrative to infer additional context beyond the given reference.
The reference may indicate category level (e.g., \textquotedblleft the
horses\textquotedblright ), or an instance appearing throughout the
entire the video (e.g., \textquotedblleft brown horse standing still\textquotedblright ),
or one appearing only in a specific segment (e.g., \textquotedblleft small
horse moving around\textquotedblright ). Based on this, it selects
an appropriate plan, enabling both efficient resource usage and case-specific
optimization. (Refer to supplementary material for additional analysis
demonstrating its advantage over rigid baselines.)

Another advantage of our Planner is its ability to refine and consolidate
vague or underspecified queries. For example, an initial input like
\textquotedblleft the horse standing still\textquotedblright{} can
be reformulated using contextual cues from the narrative into a more
precise description, such as \textquotedblleft the adult brown horse
standing still.\textquotedblright{} While this query consolidation
task is trivial for LLMs given their advanced text comprehension capabilities,
it greatly facilitates the precision of the steps that follow.

\begin{figure*}[t]
\centering{}\includegraphics[width=0.98\textwidth]{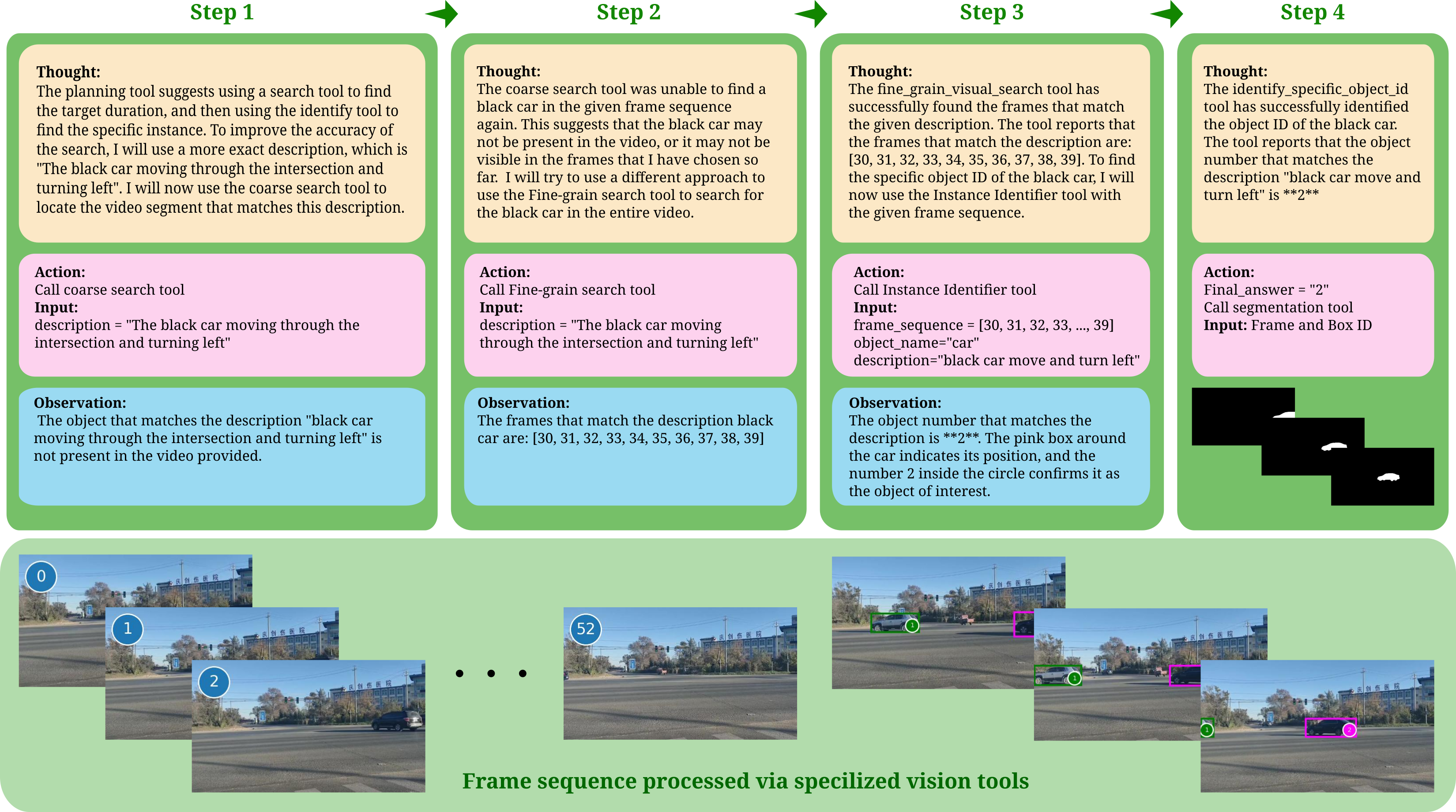}\caption{\emph{Top:} A multi-step reasoning process for the query \emph{\textquotedblleft Black
car move and turn left\textquotedblright}. The agent interprets multimodal
cues by sequentially invoking tools. Each step consists of three phases:
Thought, Action, and Observation. The process continues until a satisfactory
solution is reached. \emph{Bottom:} Specialized tools overlay the
frame sequence with temporal indices and object annotations to construct
effective prompts for the vision-language model.\label{fig:flow}}
\end{figure*}

\subsection{Multi-step Reasoning Process}

The initial plan provides a starting point for reasoning and the actual
solution is refined through tool interactions in our agentic system.
At the start, $\ModelName$ reads the system prompt, containing task
instructions and tool descriptions following specific format specifications
(see Supp.~). We then apply in-context learning \cite{wies2023learnability}
using a few reasoning examples to guide the agent before it independently
tackles real queries.

The $\ModelName$ consists of an LLM and a set of specialized tools.
The Multi-step Reasoning Process operates as an iterative interaction
between LLM and tools. At each step, we follow ReAct \cite{yao2023react}
to prompt the decision-making process, progressing through three distinct
phases: THOUGHT, ACTION, and OBSERVATION. In the THOUGHT phase, the
agent determines the next action. If tool use is required, the agent
formats input as specified in the system prompt. It then enters the
ACTION phase, generating Python code for tool calling, which is executed
externally in a separated process. The LLM relies only on the textual
responses returned by the tools for its reasoning and does not intervene
in the code execution itself. If the tool fails or returns an error,
log messages are fed back to the LLM, allowing it to adjust the plan
accordingly. This mechanism offers greater flexibility and robustness
compared to previous non-agent approaches. The $\ModelName$ receives
the tool response in OBSERVATION phase and incorporates it in the
next THOUGHT phase. The interaction continues until reaching a solution
or the step limit. If the maximum number of steps is reached, the
system defaults to the intuitive strategy: performing frame selection
followed by box selection. An example of the Multi-step Reasoning
Process is shown in the upper part of Figure \ref{fig:flow}.

\subsection{Specialized Toolset}

We develop a minimal yet sufficient toolset tailored for the task.
Each tool executes code deterministically, with input and output formats
strictly adhering to a predefined structure. Specifically, the toolset
includes:

\textbf{Audio Processing: }This tool identifies the source of sounds
from a given audio input. We use the BEATs model \cite{pmlr-v202-chen23ag}
to classify the input audio into sound sources. The five highest-scoring
categories are selected to form a candidate list.

\textbf{Temporal Search:} The search tool identifies the video segment
that best matches a given description. It takes a sequence of video
frames and a textual query as input. To convey temporal order, each
frame is overlaid with its frame number (see lower part of Figure
\ref{fig:flow}). The sampled frames are passed to the Qwen2.5VL model
along with a prompt to identify the frame sequence that best matches
the description, enabling retrieval of the most relevant segment.
We implement two variants: coarse search, which sparsely samples frames
across the entire video to capture long-duration events, and fine-grained
search, which divides the video into chunks and densely samples frames
to detect short-duration actions that coarse search may miss.

\textbf{Instance Identifier: }The Identifier tool is designed to locate
the target instance\textquoteright s ID based on a given frame sequence,
textual description, and object category. Unlike the search tool,
its objective is instance-level identification. This tool first detects
all candidates matching the specified category and overlays bounding
boxes and IDs on the frames (see lower part of Figure  \ref{fig:flow}).
The annotated sequence is fed to Qwen2.5VL with a textual prompt to
identify the object ID that best matches the description, enabling
selection of the most relevant instance.

\textbf{Video Object Segmentation and Tracking:} Given the frame index
and object ID derived from the Multi-step Reasoning Process, we use
SAM-2 \cite{ravi2024sam} to segment the target instance and propagate
the mask forward and backward through entire video.

\section{Experiments\label{sec:Experiments}}

\subsection{Experiment setup}

\textbf{Task:} We evaluate $\ModelName$ on two multimodal-conditioned
tasks: the RVOS task using the MeViS dataset \cite{MeViS}, and the
Ref-AVS task \cite{wang2024ref}, which additionally includes audio
modality in addition to textual description. Performance is measured
using the standard $\mathcal{J\text{\&}F}$ metric.

\textbf{Components of $\ModelName$ system:} We use Llama 3.3 70B
Instruct \cite{grattafiori2024llama} for both reasoning and action
generation. The toolset consists of specialized foundation models:
BEATs \cite{pmlr-v202-chen23ag} for audio classification, GroundingDINO
\cite{ren2024grounded} for text-guided object detection, SAM-2 \cite{ravi2024sam}
for video segmentation, and Qwen2.5VL \cite{wang2024qwen2} used in
the Search and Identifier tools.

The LLM is configured to halt generation upon reaching the Observation
tag, allowing it to wait for the result from tool execution. Once
the tool completes its operation, the response is appended to the
prompt, allowing the LLM to continue the reasoning process.

\subsection{Main results}

\begin{table}
\begin{centering}
\begin{tabular}{>{\raggedright}m{0.46\columnwidth}|>{\raggedright}m{0.25\columnwidth}|>{\centering}m{0.12\columnwidth}}
\hline 
Method\hspace*{0.6cm} & Reference & mIOU\tabularnewline
\hline 
\hline 
\multicolumn{3}{c}{\emph{Fully-supervised Fine-tuning}}\tabularnewline
\hline 
DsHmp \cite{DsHmp}\hspace*{0.4cm} & CVPR24 & 46.4\tabularnewline
\hline 
\hline 
\multicolumn{3}{c}{\emph{Training-free}}\tabularnewline
\hline 
G-L + SAM 2 \cite{yu2023zero} & CVPR23 & 23.7\tabularnewline
G-L (SAM) + SAM 2 \cite{yu2023zero} & CVPR23 & 30.5\tabularnewline
Grounded-SAM 2 \cite{ren2024grounded} & ICCV23{\footnotesize{} }{\scriptsize{}(Demo)} & 38.9\tabularnewline
AL-Ref-SAM 2 \cite{huang2025unleashing} & AAAI25 & 42.8\tabularnewline
\rowcolor{LimeGreen!40}$\ModelName$\hspace*{0.4cm} & \_ & 46.1\tabularnewline
\hline 
\end{tabular}
\par\end{centering}
\caption{Comparison with other methods on the MeViS dataset.\label{tab:main_result_mevis}}
\end{table}

\begin{table}
\begin{centering}
\begin{tabular}{>{\raggedright}m{0.46\columnwidth}|>{\raggedright}m{0.25\columnwidth}|>{\centering}m{0.12\columnwidth}}
\hline 
Method\hspace*{0.6cm} & Reference & mIOU\tabularnewline
\hline 
\hline 
R2VOS \cite{li2023robust} \hspace*{0.4cm} & ICCV23 & 25.01\tabularnewline
\hline 
AVSegFormer \cite{gao2024avsegformer} \hspace*{0.4cm} & AAAI24 & 33.47\tabularnewline
\hline 
GAVS \cite{wang2024prompting} \hspace*{0.4cm} & AAAI24 & 28.93\tabularnewline
\hline 
EEMC \cite{wang2024ref}\hspace*{0.4cm} & ECCV24 & 34.20\tabularnewline
\hline 
\rowcolor{LimeGreen!40}$\ModelName$\hspace*{0.4cm} & \_ & 36.26\tabularnewline
\hline 
\end{tabular}
\par\end{centering}
\caption{Comparison with other methods on the $\protect\Task$ dataset.\label{tab:main_result}}
\end{table}

Table \ref{tab:main_result_mevis} compares $\ModelName$ with other
methods on the MeViS dataset. G-L + SAM 2 \cite{yu2023zero} and Grounded-SAM
2 \cite{ren2024grounded} generate an initial mask for the first frame
and use it to prompt SAM 2. AL-Ref-SAM 2 improves this by adding a
two-step pipeline of frame and box selection, but its rigid structure
limits adaptability. In contrast, $\ModelName$ distinguishes itself
with greater flexibility, enabling case-specific planning and dynamic
reasoning. Our system constructs tailored workflows that adapt to
each scenario, yielding contextually optimized traces. Results show
that $\ModelName$ outperforms other training-free methods and is
comparable to the state-of-the-art fully supervised DsHmp \cite{DsHmp}.

We compare $\ModelName$ with state-of-the-art methods on $\Task$
in Table \ref{tab:main_result}. R2VOS \cite{li2023robust} targets
RVOS task (video + text), while AVSegFormer \cite{gao2024avsegformer}
and GAVS \cite{wang2024prompting} focus on AVS (video + audio). Wang
et al. \cite{wang2024ref} adapts these models to handle both text
and audio and introduce EEMC, the first method for Ref-AVS. All of
those methods build multimodal representations through supervised
training. In contrast, $\ModelName$ is training-free, interleaving
reasoning and action generation via LLM-guided workflows. The clear
performance gains demonstrate the strength of our agentic approach
for multimodal understanding.

\section{Conclusion}

In this work, we propose $\ModelName$, a training-free agent for
multimodal-conditioned video object segmentation. Unlike conventional
methods, $\ModelName$ leverages LLMs to generate step-by-step, case-specific
solutions through iterative reasoning. At each step, it invokes specialized
tools for low-level tasks, enabling effective and flexible execution.
The improved performance over prior approaches highlights the strength
of our agentic approach.

{\small{}\bibliographystyle{ieeenat_fullname}
\bibliography{main}
 }{\small\par}
\end{document}